(076)

# Large Language Models to Enhance Multi-task Drone Operations in Simulated Environments

Yizhan Feng [1,4], Hichem Snoussi [1,4], Jing Teng [2,] Abel Cherouat [3,4] and Tian Wang [5]

[1] UR-LIST3N, University of Technology of Troyes, Troyes, France
[2] Institute of Artificial Intelligence, North China Electric Power University, Beijing, China
[3] UR-GAMMA3, University of Technology of Troyes, Troyes, France
[4] EUt+, Data Science Lab, European Union
[5] Institute of Artificial Intelligence, Beihang University, Beijing, China
E-mail: hichem.snoussi@utt.fr

**Summary:** Benefiting from the rapid advancements in large language models (LLMs), human-drone interaction has reached unprecedented opportunities. In this paper, we propose a method that integrates a fine-tuned CodeT5 model with the Unreal Engine-based AirSim drone simulator to efficiently execute multi-task operations using natural language commands. This approach enables users to interact with simulated drones through prompts or command descriptions, allowing them to easily access and control the drone's status, significantly lowering the operational threshold. In the AirSim simulator, we can flexibly construct visually realistic dynamic environments to simulate drone applications in complex scenarios. By combining a large dataset of <natural language, program code> command-execution pairs generated by ChatGPT with developer-written drone code as training data, we fine-tune the CodeT5 to achieve automated translation from natural language to executable code for drone tasks. Experimental results demonstrate that the proposed method exhibits superior task execution efficiency and command understanding capabilities in simulated environments. In the future, we plan to extend the model's functionality in a modular manner, enhancing its adaptability to complex scenarios and driving the application of drone technologies in real-world environments.

## 1. Introduction

In recent years, drones have increasingly integrated into various aspects of daily life, finding widespread applications in fields such as environmental monitoring, communication, search and rescue, package delivery, and wireless network provisioning. While these foundational achievements have advanced the integration of artificial intelligence (AI) with unmanned aerial vehicles (UAVs), the development of a general-purpose drone system capable of executing multi-task operations remains a significant challenge [2]. Such a system requires a profound understanding of real-world physics, environmental dynamics, and the physical actions needed for task execution.

Fortunately, the rapid advancements in natural language processing (NLP) have driven the emergence of large language models (LLMs), which have demonstrated remarkable proficiency in understanding, generating, and translating human-like data. These capabilities stem from extensive training on large-scale datasets, equipping LLMs with robust general knowledge and reasoning skills. A notable example is OpenAI's ChatGPT, which has shown exceptional performance in text generation, machine translation, and code synthesis through interactive user engagement. Extending the capabilities of LLMs to the domain of drones holds revolutionary potential.

For applications requiring a combination of natural language and domain-specific terminology, LLMs have proven to be highly effective when fine-tuned on specialized datasets. These models excel in tasks demanding advanced comprehension, such as natural language-driven interactions with computational systems to generate images, program code, or structured documents. Inspired by the ability of LLMs to automate the code generation process – spanning tasks like code completion, code translation, and program synthesis [3] – we aim to leverage pre-training techniques on domain-specific drone code corpora to enable the automation of code generation for UAV tasks. This approach aspires to democratize drone operations, empowering non-specialists to execute complex tasks without requiring technical expertise or programming skills.

Vemprala, et al. [4] proposed a comprehensive framework for applying ChatGPT to robotic tasks, integrating multiple interaction modalities, including natural language dialogue and code prompting. This work systematically explores the application of ChatGPT in the field of robotics, providing in-depth analysis from theory to practice. In the domain of UAVs, the paper introduces the AirSim-ChatGPT simulation tool, based on the Microsoft AirSim platform, which provides a simple example environment for drone navigation. However, the tool has limitations in that it only allows the high-level functions of ChatGPT to be accessed in a black-box manner. Furthermore, since the commercial version of ChatGPT is not open-source, it cannot be fine-tuned with domain-specific data, significantly limiting its professional applicability in the UAV field. To address this limitation, we focus on using open-source, adjustable language models, such as CodeT5 [5], and fine-tune it using a training corpus consisting of a large number of <natural language, program code> command pairs generated by ChatGPT and UAV developer code. This approach enhances the model's controllability and transferability, whether for incremental data training or application to UAVs of different brands and models. Additionally, while Vemprala, et al [4] work primarily serves as a starting point for applying ChatGPT in drone scenarios, their proposed application is limited to a single task of industrial wind turbine inspection. In contrast, we have expanded on this foundation by incorporating a broader range of application scenarios and functional commands, further enhancing the system's utility and flexibility.

Tazir, et al [6] proposed a method that integrates OpenAI's ChatGPT with the PX4/Gazebo simulator, aimed at controlling drones within the simulator through natural language commands. This integrated approach provides users with a realistic and safe environment for drone control training, while also supporting the execution of XML code within the PX4/Gazebo environment, facilitating the testing and optimization of control algorithms. However, this method shares similar domain limitations with the approach proposed by Vemprala et al. [4]. Additionally, the PX4/Gazebo environment is comparatively rudimentary in relation to AirSim, lacking the capability to effectively replicate the realism of real-world scenarios, which limits its applicability in real-world settings. AirSim, as a drone simulator based on the Unreal Engine, offers more advanced features. In addition to being compatible with PX4, AirSim supports a wider variety of drone types. Its distinguishing characteristics include highly realistic physical and visual simulations, the ability to dynamically integrate Unity game plugins, and seamless integration into any Unreal environment. Moreover, the AirSim platform boasts a wealth of modules and environments contributed by developers, allowing users to customize and invoke these modules to enrich the simulation environment. This flexibility enables the creation of complex scenarios that may exceed the capabilities of real-world environments, allowing for comprehensive drone performance testing without the constraints of real-world costs and limitations [7].

## 2. Methodology

We developed a system that enables users to control drones using natural language commands or prompts. The construction of the fine-tuning training dataset consists of two components. The first component focuses on a general set of simple commands, maintained as a JSON file mapping natural language commands to Python code snippets. Leveraging the generative capabilities of ChatGPT, we generated a large number of <natural language, program code> command-execution pairs to cover basic drone operations, such as flight control, state configuration and retrieval, and image acquisition. The second component addresses complex task operations, including object recognition, autonomous navigation, and dynamic tracking. For this, we employed developer-provided code as training data, dynamically maintaining a <task prompt, program code> to serve as the complex functionality set. As illustrated in Fig. 1, the overall workflow of the proposed methodology is effectively presented, providing a clear and intuitive representation of the process.

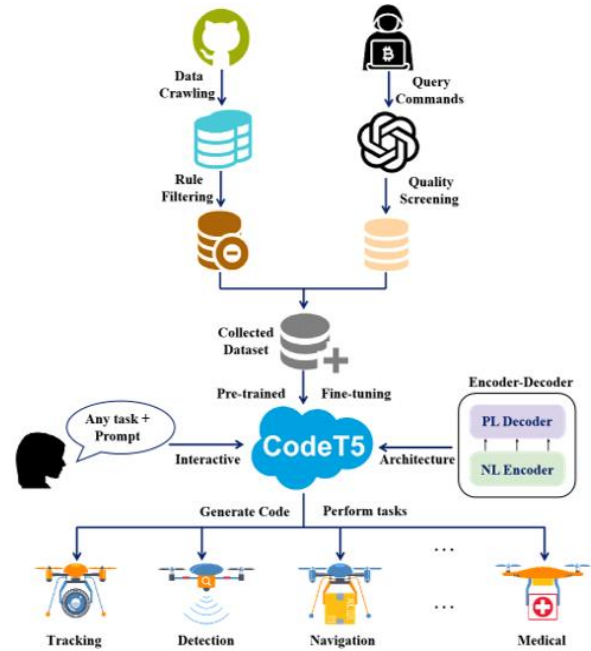

**Fig. 1.** Overview of the proposed methodological framework. Our research concentrates on the task of mapping natural language (NL) inputs to corresponding programming language (PL) outputs. We initiate this process by constructing a proxy dataset, followed by the generation of query-result pairs derived from API documentation or task-specific knowledge. Subsequently, we train the CodeT5 model using this curated dataset, which consists of query-result pairs designed to facilitate the learning of NL-to-PL mappings.

We fine-tuned the open-source LLM CodeT5 [5] leveraging its T5 encoder-decoder architecture and extensive pre-training on code datasets. CodeT5 demonstrates exceptional performance in code translation and syntactic correctness, making it highly suitable for drone-specific data. After fine-tuning, CodeT5 generates executable code directly from input natural language commands or prompts, which can be seamlessly integrated into our Python framework for deployment in the AirSim simulation environment [8].

Unlike ChatGPT, which is designed for detailed and verbose conversational interactions, our system emphasizes concise and efficient code generation to prioritize rapid command execution. This design reduces response latency and avoids unnecessary drone idling during operation, making it more effective for real-world applications where timely execution is critical.

The system integrates a Python-based conversational agent with a C++ interface for the AirSim simulator. After establishing an asynchronous connection, the agent processes user commands via the fine-tuned CodeT5 model, generating executable code displayed for user confirmation. Once validated, the system seamlessly transmits the command to the simulated drone in AirSim, ensuring real-time and accurate execution within the simulation environment.

## 3. Experiments

To comprehensively assess the accuracy of command generation, we designed a series of test scenarios that encompass a variety of natural language inputs representing different drone actions. These input scenarios include common drone operations such as basic flight control, state configuration and retrieval, and image acquisition. Each test scenario is based on task requirements from real-world applications, ensuring that they accurately reflect the impact of natural language commands on the drone control system.

In the evaluation process, we focused on whether the generated commands accurately executed the intended tasks and assessed their compliance by comparing them with the established command structure. Specifically, the evaluation criteria include the syntactic correctness of the command, the effectiveness of the generated results, and their consistency with predefined command formats. Additionally, to ensure the objectivity of the evaluation, we further validated the system's reliability by comparing the execution results of manually written standard commands with those of the generated commands.

Fig. 2 illustrates the successful implementation of the control process. The figure depicts the entire process from the user inputting a natural language command to the drone's response and execution of the task, clearly presenting the seamless integration of command issuance, code generation, and command execution. Based on the natural language instructions provided through the system's interactive prompts (as illustrated in the blue section of the figure), the trained model translates these instructions into corresponding executable code lines (as shown in the red section of the figure). Upon user confirmation, the generated instructions are transmitted to the drone within the simulation environment for execution. Following execution, the system returns relevant outputs, including images, status information, and video data.

## 4. Conclusion

We present a system that uses fine-tuned CodeT5 to enable natural language-driven control of drones within a simulation environment. By creating a specialized dataset of natural language-command and program code pairs, we fine-tuned CodeT5 to handle both simple and complex drone tasks, ensuring adaptability across different UAV platforms. The system prioritizes efficiency, reducing latency for real-time task execution and minimizing drone idle time. Through integration with the AirSim simulation environment, the system benefits from advanced physical simulations for realistic drone control testing. The Python-based conversational agent, combined with a C++ interface, enables seamless command generation and execution, enhancing system performance.

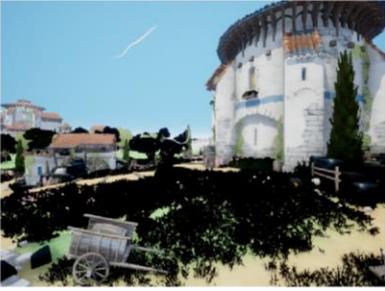

**Fig. 2.** Example of commands and executions.

Our approach demonstrates progress in AI-driven drone control systems and sets the stage for further research into more complex, real-world applications of natural language interfaces for UAV operations. Future work will focus on expanding task categories and applying the system to real-world UAVs.

## Acknowledgements

This work has been partially funded by BPI DreamScanner project.